\documentclass[10pt,twocolumn,letterpaper]{article}

\usepackage{cvpr}
\usepackage{epsfig}
\usepackage{amssymb}
\usepackage{times}
\usepackage{helvet}
\usepackage{courier}
\usepackage{url}
\usepackage{graphicx}
\usepackage{subfigure} 
\usepackage{tikz}
\usepackage{pgfplots}
\usepackage{graphicx} 
\usepackage{amsmath} 
\usepackage{verbatim}

\usepackage[pagebackref=true,breaklinks=true,letterpaper=true,colorlinks,bookmarks=false]{hyperref}

\newcommand{\oma}[1]{{\color{red} \bf [OMA: #1]}}

\cvprfinalcopy 

\def\cvprPaperID{1580} 

\ifcvprfinal\fi

\begin{document}
\pagenumbering{gobble}
\title{Context Embedding Networks}
\author{Kun Ho Kim\hspace{20pt}Oisin Mac Aodha\hspace{20pt}Pietro Perona\\
California Institute of Technology}
\maketitle


\begin{abstract}
Low dimensional embeddings that capture the main variations of interest in collections of data are important for many applications.
One way to construct these embeddings is to acquire estimates of similarity from the crowd.
Similarity is a multi-dimensional concept that varies from individual to individual.
However, existing models for learning crowd embeddings typically make simplifying assumptions such as all individuals estimate similarity using the same criteria, the list of criteria is known in advance, or that the crowd workers are not influenced by the data that they see.

To overcome these limitations we introduce Context Embedding Networks (CENs). 
In addition to learning interpretable embeddings from images, CENs also model worker biases for different attributes along with the visual context i.e.\ the attributes highlighted by a set of images.
Experiments on three noisy crowd annotated datasets show that modeling both worker bias and visual context results in more interpretable embeddings compared to existing approaches. 

\end{abstract}

\vspace{-10pt}
\section{Introduction}\label{sec:introduction}

Large annotated datasets are a vital ingredient for training automated classification and inference systems. 
Labeling these datasets has been made possible by crowdsourcing services, which enable the purchasing of annotations from crowd workers. 
Unfortunately fine-grained categorization is very challenging for untrained workers. 
The alternative, obtaining annotations from experts, is equally impractical due to the fact that for many domains experts are few~\cite{vanHornCVPR2015}.
Instead of obtaining semantic fine-grained category-level labels, one can ask workers to label images in terms of their similarities and differences. 
This is intuitively much easier for untrained workers because it requires the comparison of images, a task that humans are naturally good at.
This approach, however, presents its own challenges: 1) different workers may use different criteria when estimating the similarity between pairs of images, and 2) workers may be influenced by the set of images that they see when making their decisions i.e. `context'. 

\begin{figure}[t!]
\centering
\includegraphics[width=0.40\textwidth]{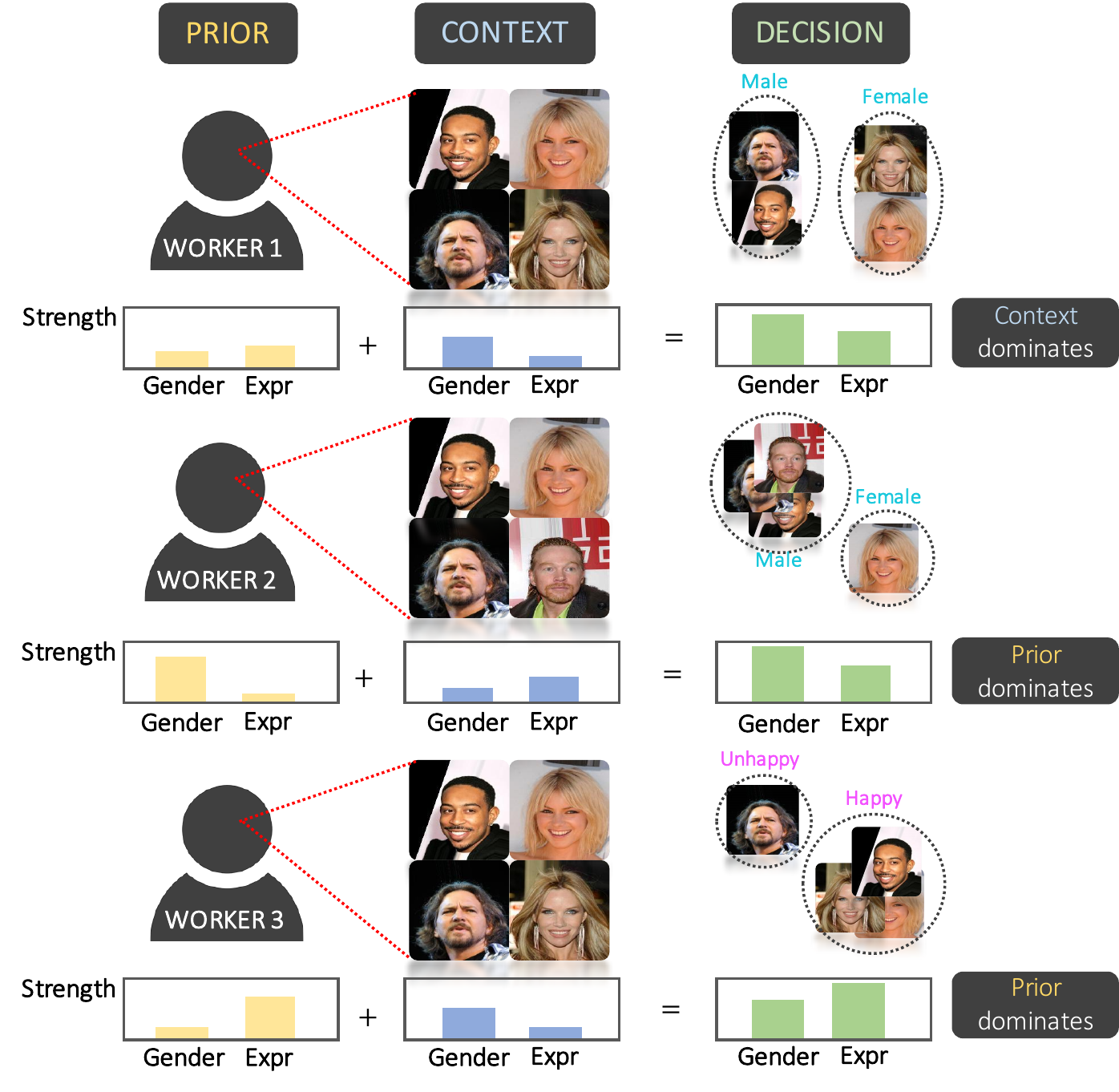}
\vspace{-5pt}
\caption{{\bf Context influences similarity estimates.} We hypothesize that estimating similarity according to a particular visual attribute is influenced by a combination of innate biases and the context in which these decisions are made. Compared to worker 1, worker 2 has a strong prior bias towards using the gender attribute. Influenced by the context of the images worker 1 also groups based on gender. Worker 3 sees the same context as worker 1 but ultimately groups based on expression due to prior bias.} 
\label{fig:attsim}
\vspace{-15pt}
\end{figure}

In Fig. \ref{fig:attsim} we see an example of three different crowd workers estimating similarity by clustering a collection of images. 
The workers' decision for which visual attribute they use to compare the images can be explained by two factors:
1) The workers have an innate preference towards certain attributes based on their past experiences and 2) the set of related images that a worker observes biases them towards certain attributes. 
We call this first bias the \emph{worker prior} and the second bias the \emph{context}.
Our hypothesis is that different sets of images highlight different visual attributes to the workers.
The majority of existing work often assumes that all workers behave in the same way \cite{van2012stochastic}, the list of  attributes are specified in advance \cite{veitconditional}, or in addition to similarity estimates, workers also indicate which attributes they used to make their decision \cite{tian2017learning}.

We introduce Context Embedding Networks (CENs), an efficient end-to-end model that learns interpretable, low dimensional, image embeddings that respect the varied similarity estimates provided by different crowd workers. 
Our contributions are: 
1) A flexible model that produces an embedding for a set of input images. This is achieved by modeling worker bias and image context i.e.\ the degree to which each worker is influenced by the attributes present in a given set of images.
2) An empirical evaluation on annotations from real crowd workers showing that CENs outperform existing approaches, producing interpretable, disentangled, low-dimensional feature spaces. 
\begin{figure*}[h]
\centering
\includegraphics[width=0.9\textwidth]{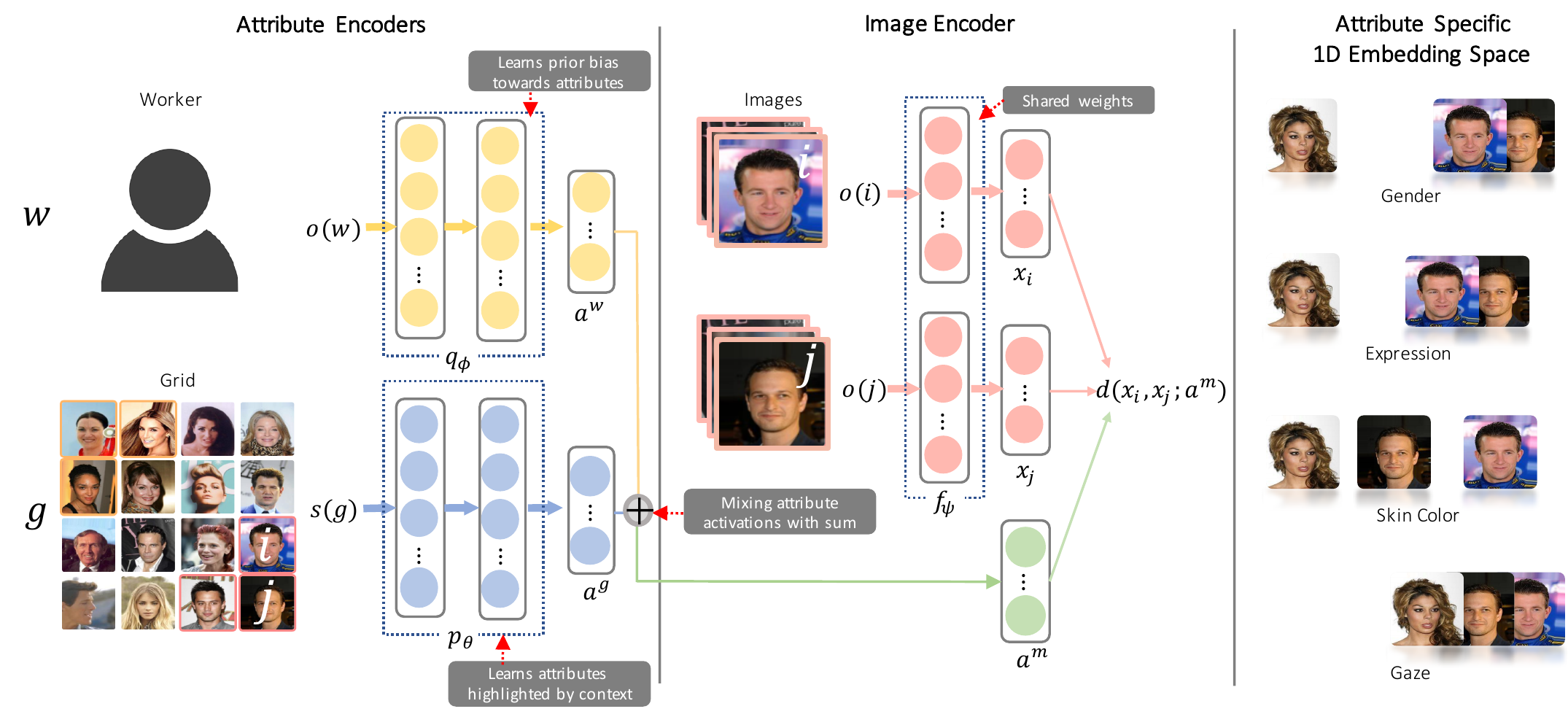}
\vspace{-10pt}
\caption{{\bf Context Embedding Networks} are composed of three neural networks that are trained jointly. (Top-Left) A worker encoder network models workers' annotation behavior and (Bottom-Left) a context encoder network models the attributes highlighted by a particular set of images. Jointly, these networks are referred to as the attribute encoders and are used to weight the embeddings produced by the image encoder network (Center). (Right) Our final embedding respects similarity estimates from each worker in the same low dimensional space where each dimension corresponds to a different visual attribute.}
\label{fig:concept}
\vspace{-10pt}
\end{figure*}
\section{Related Work}\label{sec:related}

\textbf{Learning Embeddings}
The goal of embedding algorithms is to learn a low dimensional representation of a collection of objects (e.g.\ images), such that objects that are ``close'' in the potentially high dimensional input space are also ``close'' in the embedding space. 
Embeddings are useful for a large number of tasks from face recognition \cite{schroff2015facenet} to estimating the clinical similarities between medical patients \cite{zhu2016measuring}.
They can be learned from pre-defined feature vectors representing the input objects \cite{maaten2008visualizing}, from similarity estimates obtained from the crowd \cite{tamuz2011adaptively,van2012stochastic}, or a combination \cite{wilber2015learning}. 
Crowdsourced annotations can come in the form of pairwise \cite{gomes2011crowdclustering} or relative similarity estimates \cite{tamuz2011adaptively,vinayak2016crowdsourced}. 
Presenting workers with sets of images, as opposed to pairs or triplets, is an efficient way of acquiring estimates of similarity \cite{gomes2011crowdclustering,wah2014similarity,wilber2014cost}.
Another approach is to learn a function that can extract meaningful features from the raw input data by training on similarity labels e.g. \cite{bromley1993signature,veitconditional}. 
This has the advantage of being able to also embed objects not observed at training time. 

\textbf{Different Notions of Similarity}
A limitation of the above methods is that they typically assume that objects are compared using a single similarity criteria.
Given a pair or triplet of images, one estimate of similarity may be valid for one visual attribute, or trait, but invalid for another.
For example, in Fig. \ref{fig:attsim} comparing faces according to gender or expression will result in a different grouping. 
In practice, workers may use different criteria unless they are specifically told which attribute to use. 
To overcome this limitation there is a body of work that attempts to learn embeddings where alternative notions of similarity are represented in the embedding space. 
One common approach is to instruct the workers to provide additional information regarding the attribute they used when making their decision.
This information can come in multiple forms such as category labels \cite{veitconditional}, user provided text descriptions \cite{tian2017learning}, or part and correspondence annotations \cite{maji2014part}.

Similar to \cite{wah2014similarity}, \cite{amid2015multiview} propose a model inspired by \cite{van2012stochastic} that produces a separate embedding for each similarity criteria instead of learning a single embedding that tries to satisfy all constraints.
In contrast, \cite{veitconditional} learn a unified embedding where alternative notions of similarity are extracted by masking different dimensions in this space.  
However, the visual attribute used for each similarity estimate is assumed to be known.
\cite{wang2016contextual} also learn a weighted feature representation of the input examples but require category level labels in order to learn cross-category attributes.
Their model learns a different weight vector for each triplet, resulting in a large number of parameters. 
\cite{tian2017learning} propose a generative model for learning attributes from the crowd where workers are instructed to specify an attribute of interest via a text box and then perform similarity estimates for a set of query images based on these pre-defined attributes.
The majority of these methods assume that extra information, in addition to the pairwise or triplet labels, are available to the model.
We instead make use of the context information that is present in the set of images that we show to our crowd workers.

\textbf{Modeling the Crowd}
Crowdsourcing annotations is an effective way of gathering a large amount of labeled data \cite{kovashka2016crowdsourcing}.
One difficulty that arises when using such annotations is that they can be noisy, as workers behave differently. 
One solution to this problem is to model the ability and biases of each worker to resolve better quality annotations \cite{whitehill2009whose,welinder2010multidimensional,Branson_2017_CVPR}. 
Specific to clustering, \cite{gomes2011crowdclustering} propose a Bayesian model of how workers cluster data from noisy pairwise annotations.
To efficiently gather a large number of labels, workers are presented with successive grids of images and are asked to cluster the images into multiple different groups. 
By modeling individual workers as linear classifiers in an embedding space they allow for different worker biases. 
However, they assume that workers are consistent in the criteria they use when making their decisions and that it does not change over time.
Our approach also learns individual worker models while also making use of the strong context information provided by the image grid. 

\textbf{Attribute Discovery}
Low dimensional, attribute based, representations of images have the benefit of being more interpretable than raw pixel information \cite{farhadi2009describing, feris2017visual}.
In addition to providing semantically understandable descriptions of images, they can also be used for applications such as zero shot learning \cite{lampert2009learning}.
Attributes can be discovered by various means, from mining noisy web images and their associated text descriptions \cite{berg2010automatic} to crowdsourcing \cite{parikh2011relative}.
In this work, while we do not explicitly aim to produce `nameable' attributes, we qualitatively observe that the embeddings that our model produces are often disentangled along the embedding dimensions.


\section{Methods} \label{sec:methods}

We crowdsource the task of image similarity estimation for a dataset containing $N$ images referenced by $i, j = 1, ..., N$. 
Each crowd worker $w = 1, ..., W$, is presented with an image grid $g = 1, ..., G$, displaying a collection of images $\{i_{g}\}$ which they group into as many categories as they wish \cite{gomes2011crowdclustering}. 
A grid of $S$ items results in $(S^2 - S)/2$ pairwise labels, e.g. a single grid of 24 items produces the same number of annotations as 276 individual pairs.
Across grids, real workers are often inconsistent with the attributes they use to cluster and the number of clusters they create. 
A pair of images $(i_{g}, j_{g})$ shown in the same grid, $g$, clustered by worker $w$ is assigned a positive label $l = 1$ if they are grouped together and $l = 0$ otherwise.
This results in a training set of pairwise similarity labels
\begin{equation}\label{eq:dataset}
\mathcal{D} = \{(w, g, i_{g}, j_{g}, l) | g = 1, ..., G\}.
\end{equation}

\subsection{Context Embedding Network (CEN)}
Here we present out CEN model and define the loss function used to train it. This involves joint training of three networks which model workers, grid context, and image embedding respectively, see Fig. \ref{fig:concept}. The first two networks are referred to as \textit{attribute encoders} while the third is the \textit{image encoder}. 

\subsubsection{Worker Encoder} For the workers we define an attribute encoder network $q_{\phi}$ which takes as input a one-hot encoding $(o(\cdot))$ of worker $w$ and outputs a $K$ dimensional worker, \textit{attribute activation}, vector $a^{w} = q_{\phi}(o(w)) = [a^{w}_{1}, ..., a^{w}_{K}]$.
Each $a^{w}_{k}$, for $k = 1, ..., K$, represents the degree of prior bias towards attribute $k$ for worker $w$. 
Once the network is trained, the output attribute activation vector models the worker's prior preferences for each visual attribute. 
For example, a heavily biased worker that only attends to a single attribute $k^{*}$ should have high activation for that particular attribute dimension $a^{w}_{k^{*}}$. 
On the other hand, a worker that does not have a strong preference for any particular attribute will have weak attribute activations in all $K$ dimensions and may be more influenced by the grid context.

\subsubsection{Context Encoder} For an image grid containing $S$ images, we define a context encoder network $p_{\theta}$ that takes as input a $S$-hot encoding ($s(\cdot)$) of the grid $g$ and outputs a $K$ dimensional grid \textit{attribute activation} vector $a^{g} = p_{\theta}(s(g)) = [a^{g}_{1}, ..., a^{g}_{K}]$. 
Each $a^{g}_{k}$ for $k = 1, ..., K$ represents the degree of visual prominence of attribute $k$ for grid $g$.
Once the network is trained, the grid attribute activation dimensions with high values should correspond to the most salient visual attributes highlighted by the input grid. Intuitively, \emph{attribute variance} in the collection of images should influence which attributes are more noticeable to workers. For instance, a collection of images that is similar along all other attributes except one $k^{*}$ should have a peak activation at $a^{g}_{k^{*}}$. On the other hand, if the image set varies along many different attributes, $a^{g}$ should be close to uniformly distributed. 
The attribute vectors $a^{w}$ and $a^{g}$ from the worker and context encoders are combined to produce the final attribute encoder output $a^{m}$ (Fig. \ref{fig:concept} Center).

\subsubsection{Image Encoder} We seek to learn a non-linear mapping from image $i$ to a disentangled Euclidean coordinate $x_{i}$ where each dimension embeds the image into a one dimensional attribute specific subspace. 
To achieve this we use a Siamese Network architecture for the image encoder network $f_{\psi}$ with shared parameters $\psi$ that take as input a one hot encoding of image $i$ and outputs a $K$ dimensional embedding vector $x_{i} = f_{\psi}(o(i)) = [x_{i1}, ..., x_{iK}]$. Although our image embedding network learns an embedding for each input image directly, with enough data it is possible to learn a feature extractor from the raw images \cite{veitconditional}.
Similarly, we present our model in terms of a pairwise loss, but it is also possible to use a triplet loss for the image encoder. 
For brevity, from this point forward we omit the one and S-hot encoding function notation $o(\cdot), s(\cdot)$.

\subsection{Learning from the Crowd}
By ignoring worker and context information, an embedding can be learned using Siamese networks \cite{bromley1993signature}, where the contrastive training loss $L_{c}$ is defined as
\begin{equation}\label{eq:basicloss}
L_{c}(x_{i}, x_{j}) = ld(x_{i}, x_{j}) + (1 - l)\max\{0, \xi_{n} - d(x_{i}, x_{j}) \}, 
\end{equation}
where $d(x_{i}, x_{j}) = \lVert x_{i} - x_{j} \rVert_{2}$ is the $L2$ distance between image $i, j$ in embedding space.
$\xi_{n}$ is the negative margin which prevents over-expanding the embedding manifold, and $l \in \{0,1\}$ is the user provided label. 
This contrastive loss alone does not encourage the network to learn low dimensional attribute specific embeddings as it assumes that all crowd workers compare images using the same visual attributes.
To overcome this, we weight the $L2$ distance metric by the attribute activation vectors $a^{w}$ and $a^{g}$. 
We hypothesize that a worker's decision to cluster along a particular attribute depends on both their prior preferences for specific visual attributes and the context highlighted by the set of images in the grid. 
Based on this assumption, we define three variants of the distance metric weighted by the attribute activation vectors
\begin{equation}\label{eq:workerdist}
\begin{split}
d(x_{i}, x_{j}; a^{w}) &= \lVert a^{w} \cdot (x_{i} - x_{j}) \rVert_{2} \\
&= \lVert q_{\phi}(w) \cdot (f_{\psi}(i) - f_{\psi}(j)) \rVert_{2}
\end{split}
\end{equation}

\begin{equation}\label{eq:griddist}
\begin{split}
d(x_{i}, x_{j}; a^{g}) &= \lVert a^{g} \cdot (x_{i} - x_{j}) \rVert_{2} \\
&= \lVert p_{\theta}(g) \cdot (f_{\psi}(i) - f_{\psi}(j)) \rVert_{2} 
\end{split}
\end{equation}

\begin{align}\label{eq:mixeddist}
d(x_{i}, x_{j}; a^{m}) &= \lVert a^{m} \cdot (x_{i} - x_{j}) \rVert_{2} \\
&= \lVert (p_{\theta}(g) + q_{\phi}(w)) \cdot (f_{\psi}(i) - f_{\psi}(j)) \rVert_{2},\nonumber
\end{align}
where $a^{m} = a^{g} + a^{w}$ is the mixed attribute activation vector. 

After exploring different non-linear methods of mixing $a^{g}$ and $a^{m}$, we found that a simple summation sufficiently captures the relationship between the two biases.
In the experiments section, we compare the performance of the above three different models. For the model in Eq. \ref{eq:mixeddist}, biased workers should have a concentrated worker attribute activation vector $a^{w}$ which will dominate the mode of sum $a^{m} = a^{w} + a^{g}$. Alternatively, workers with weak prior preferences should have low worker attribute activations $a^{w}$ and the grid attribute activations $a^{g}$ will dictate the mode. 
Intuitively, the attribute activation vector serves as a mask which indicates the embedding dimension that should be weighted heavily in the loss e.g.\ \cite{veitconditional}. By encouraging sparsity in $a^{w}$ and $a^{g}$ along with ReLU non-linearities \cite{nair2010rectified}, we assume that grids that were clustered along one attribute will have a uni-modal $a^{m}$ while grids that were clustered on a mixture of attributes will have a multi-modal $a^{m}$ with peaks corresponding to the attribute dimensions used. 

Inspired by the dual margin contrastive loss proposed in \cite{wang2016contextual}, we include a positive margin term $\xi_{p}$ in the loss function to prevent two images from overlapping in the embedding space which could lead to over fitting. 
This ensures that images will be pushed closer only if their current embedding is separated by more than $\xi_{p}$.
We use $a$ to denote the general attribute activation vector which can be $a^{g}, a^{w}$, or $a^{m}$ depending on the model variant 
\begin{equation}\label{eq:margin}
\begin{split}
L_{c}(x_{i}, x_{j}; a) = &l\max\{0, d(x_{i}, x_{j}; a) - \xi_{p}\} + \\
&(1 - l)\max\{0, \xi_{n} - d(x_{i}, x_{j}; a) \}.
\end{split}
\end{equation}

A crowd worker's decision to group two images is an active decision while choosing not to group images together can be seen as a more passive decision. 
This can become a problem when workers group images with different levels of detail. For example, a grid of shapes containing squares, triangles, circles, and stars might be clustered into two groups, squares and non-squares, by one worker. A second worker may group the images into the four different shape types. An embedding model might incorrectly assume that a different attribute was used to separate the images, when it is in fact just a different level of granularity of `shape' that is being used by both workers.
To overcome this problem, we introduce an additional positive similarity weight $\gamma$, that captures the relative importance of the positive similarity labels compared to the dissimilarity labels 
\begin{equation}\label{eq:labelweight}
\begin{split}
L_{c}(x_{i}, x_{j}; a) = &\gamma l\max\{0, d(x_{i}, x_{j}; a) - \xi_{p}\} + \\
&(1 - l)\max\{0, \xi_{n} - d(x_{i}, x_{j}; a) \}.
\end{split}
\end{equation}
This ensures that the model can learn the high level attributes when workers cluster with different levels of detail. In the example above, although cross category labels between circles, triangles, and stars are $l = 0$, the positive labels generated within each circle, triangle, and star groups agree with the positive labels generated within the non-square group thus allowing the network to learn that the high level attribute, i.e.\ shape, used by both workers are the same. We show the impact of $\gamma$ on the performance of our CEN in the supplementary materials.

\subsection{Regularization}
We add $L1$ penalties $\lambda_{1}\lVert a \rVert_{1}$ to the attribute encoders to encourage sparsity in the attribute activation vector.
We also regularize the embedding network with a $L2$ penalty $\lambda_{2}\lVert x \rVert_{2}$ to encourage regularity in the latent space. The final loss function for our CENs is 
\begin{equation}\label{eq:loss_all}
\begin{split}
L_{CEN}(x_{i}, x_{j}; a) = &\gamma l\max\{0, d(x_{i}, x_{j}; a) - \xi_{p}\} + \\
&(1 - l)\max\{0, \xi_{n} - d(x_{i}, x_{j}) \} + \\
&\lambda_{1}\lVert a \rVert_{1} + \lambda_{2}\lVert x_{i} \rVert_{2}+ \lambda_{2}\lVert x_{j} \rVert_{2}.
\end{split}
\end{equation}
CENs require the number of dimensions $K$ as a hyper-parameter. 
However, we observe that by setting $K$ to a large number and by $L1$ regularizing $a^{w}$ and $a^{g}$, our model tends to only use a subset of the available embedding dimensions.

\setlength{\itemindent}{-.5in}

\section{Experiments} \label{sec:results}
Here we show that CENs can recover meaningful low-dimensional embeddings from noisy data.
Network architectures, training details, and hyperparameters tuning are described in the supplementary material.
We perform experiments on the following three datasets:
\newline\textbf{\textsc{CelebA}} contains images of different celebrity faces from which we select a random subset of 300 images \cite{liu2015faceattributes}. 
For this dataset we instruct workers in advance to cluster on one attribute per grid respecting four visual attributes: gender, expression, skin color, and gaze direction. 
Although we expect some workers to deviate from our instructions, having a definite ground truth set of attributes allow us to quantify the attribute retrieval accuracy. 
The CEN is unaware of the attribute selected for each grid. 
In total, 94 workers clustered 620 grids, yielding 170,000 similarity training pairs.
\newline\textbf{\textsc{Retina}} is a medical dataset comprising of fundus images of the retina belonging to patients with varying degrees of diabetic retinopathy \cite{kaggleDR}.
The images contain a number of visual indicators for the disease such as hard exudates (yellow lesions dispersed throughout the retina). 
From 66 fundus images we crop out 300 image patches. 
These patches provide a localized view that may or may not contain indicator features of the disease. 
This dataset is more challenging to discover meaningful attributes as the disease indicator features are visually subtle and the images are unfamiliar to the crowd. 
We do not provide any instructions as to the attributes the workers should use for this dataset. 
62 workers clustered 620 grids, yielding 170,000 similarity pairs. 
\newline\textbf{\textsc{Birds}} is a larger dataset composed of 1000 bird head images made up of 16 randomly selected species from \cite{wah2011caltech}. We use this dataset to demonstrate the scalability of our CENs. 252 workers clustered 3,000 grids yielding 820,000 similarity labels. 

\begin{figure}[t]
\centering
\includegraphics[width=0.4\textwidth]{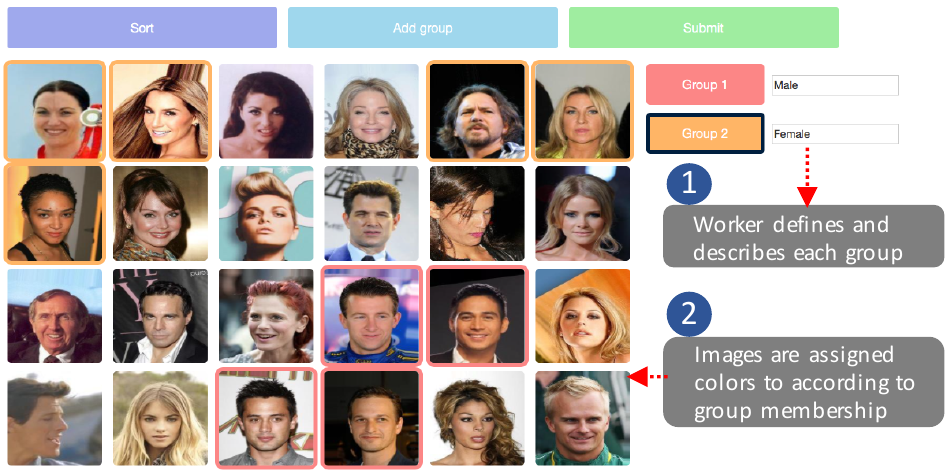}
\vspace{1pt}
\caption{{\bf Data collection GUI.} Workers group images they perceive to be visually similar by assigning them to different groups. They can create up to ten groups per grid of images.
}
\label{fig:gui}
\vspace{-15pt}
\end{figure}

\subsection{Data Collection}
We use Amazon Mechanical Turk's crowdsourcing platform to request crowd workers to cluster grids of images using the GUI shown in Fig. \ref{fig:gui}. 
Workers were presented with a $4 \times 6$ grid of images randomly sampled from the given dataset. 
Using up to ten possible groups, workers clustered images by first clicking on a group button on the right side of the page then clicking on the desired images. 
For each group they were asked to provide a short text description, used only for evaluation.
The image, cluster, and worker ids were then converted into pairwise similarity labels (Eq. \ref{eq:dataset}).
Each worker clustered a minimum of ten grids in order to receive a reward, ensuring that the worker encoder network had sufficient data to learn from. 

\subsection{Baseline Comparisons}
We compare results to four baseline methods and three variants of our model: 
\newline\textbf{Standard Siamese Network e.g. \cite{bromley1993signature}}: Assumes that all pairwise similarity labels come from the same notion of similarity, as in Eq. \ref{eq:basicloss}.
\newline\textbf{Standard Triplet Network e.g. \cite{schroff2015facenet}}: Learns embeddings given similarity labels of the form "A is more similar to B than C". 
\newline\textbf{Bayesian Crowd Clustering \cite{gomes2011crowdclustering}}: 
Workers are modeled as linear classifiers in the embedding space where both an  \textit{entangled} image embedding and individual worker models are jointly learned with variational methods. 
\newline\textbf{CSN \cite{veitconditional}}: Learns an entangled image embedding from similarity triplets which are disentangled by masks learned separately for each pre-known attribute. This baseline represents the situation where the similarity dimension used by the worker is \emph{known}.
\newline\textbf{CEN-worker encoder only}: This first variant of our model uses only worker modeling to learn attribute activations which weight the embeddings as in Eq. \ref{eq:workerdist}.
\newline\textbf{CEN-context encoder only}: 
Here we only model context information to weight the embeddings as in Eq. \ref{eq:griddist}.
\newline\textbf{CEN-mixture}: Our full model, incorporates both worker and context information to learn a network that weights the worker bias $a^{w}$ and grid context $a^{g}$ as in Eq. \ref{eq:mixeddist}.

\subsection{Unsupervised Attribute Retrieval}
\begin{figure}[t]
\centering
\includegraphics[width=0.48\textwidth]{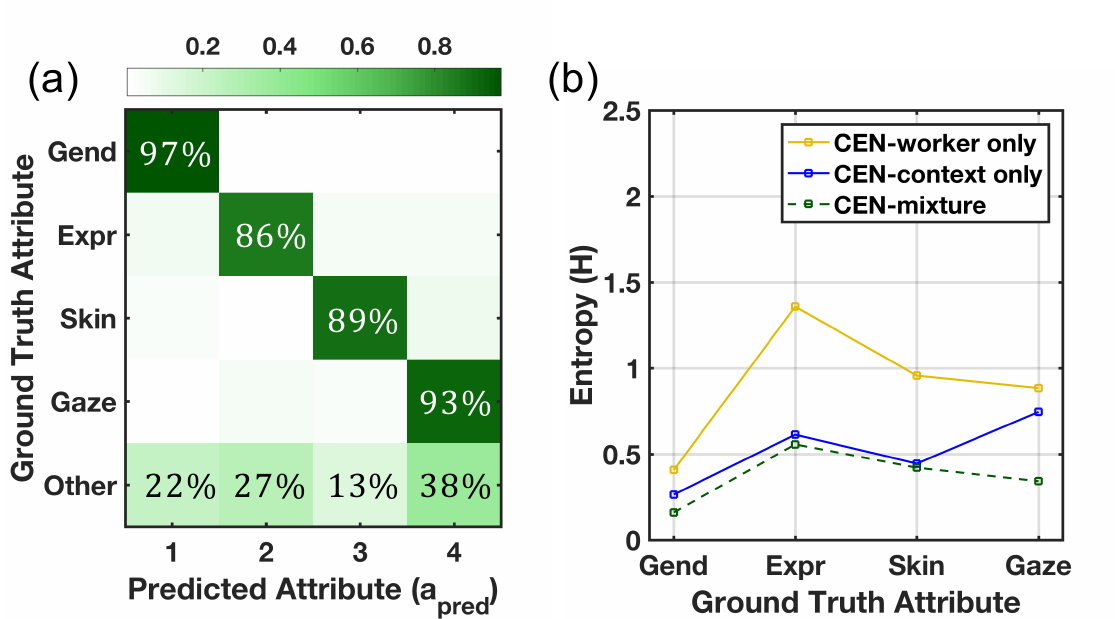}
\vspace{-10pt}
\caption{{\bf Attribute retrieval accuracy.} On the left we see the predicted embedding dimensions from the CEN-mixture model compared to the ground truth visual attributes for the \textsc{CelebA} dataset. On the right, we quantify how disentangled the learned embeddings are. Lower entropy indicates models that better capture the ground truth attributes along individual embedding dimensions.}
\label{fig:attacc}
\vspace{-10pt}
\end{figure}

First, we evaluate whether our CEN can accurately recover the four dominant attributes present in the \textsc{CelebA} dataset. For each grid $g$ clustered by worker $w$, we take the mode dimension of the attribute activation vector $a$ to be the model's prediction, $a_{pred}  = \text{argmax}_{k} a_{k}$. This is the attribute that we predict was used to cluster the set of images. Again $a$ can be $a^{w}, a^{g}$ or $a^{m}$ depending on the model variant used.
We then examine the annotations provided by workers for each set of grids that map to a different $a_{pred} \in \{1, 2, 3, 4\}$ and quantify the proportion of each attribute actually used. 
In Fig. \ref{fig:attacc}(a) we show a confusion matrix illustrating that for each worker and grid pair, the CEN-mixture model is able to accurately predict which attribute was used. The row for \textit{gender} and the first column denote the proportion of grid submissions that have $a_{pred} = 1$ out of all the submissions that were clustered along gender. For all attributes we obtain over $85\%$ attribute prediction accuracy. In Fig. \ref{fig:attacc}(b) we plot the entropy $H$ of the distribution $p$ for each row of the confusion matrix where $H_{p} = -\sum p\log{p}$.
High entropy indicates that the ground truth attributes are scattered throughout the attribute predictions and vice versa. 
The CEN-mixture model learns the most disentangled embeddings across the four ground truth attributes compared to its variants. 

\begin{figure}[h]
\centering
\includegraphics[clip,trim=0 0 0 20pt, width=0.45\textwidth]{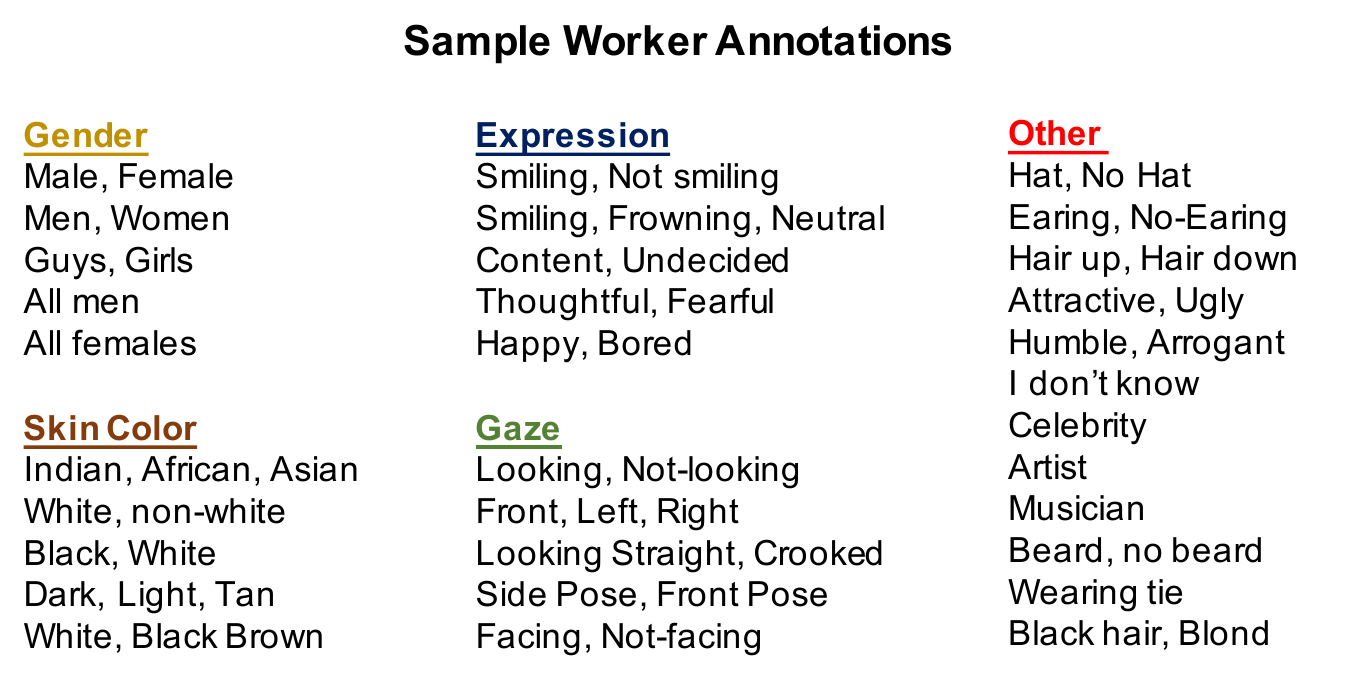}
\caption{{\bf Cluster Names.} Keywords provided by workers for \textsc{CelebA}. Colored labels indicate the manual grouping performed by us (only used for evaluation). Some workers use finer grained distinctions compared to others.}
\label{fig:sample_annot}
\vspace{-10pt}
\end{figure}

Although workers were encouraged to focus on four different attribute options for this experiment, in practice they did not abide by our instructions and the proportion of noise in the raw data is significant. For the \textsc{CelebA} dataset approximately $19.1\%$ of the HITs completed were either clustered on different attributes such as ``wearing sun glasses'' (see Fig. \ref{fig:sample_annot}) or noisy submissions where images were not separated into different groups. 
We also observed workers using different levels of detail when clustering on the same attribute. For example, for the \textit{gaze} attribute some workers labeled ``looking left'', ``looking right'', etc. To demonstrate our model's robustness, we perform all of our experiments on this raw data without filtering out annotation noise.  
For evaluation of the worker model learned by the worker encoder, refer to the supplementary material. 

\begin{figure*}[h]
\centering
\includegraphics[width=0.9\textwidth]{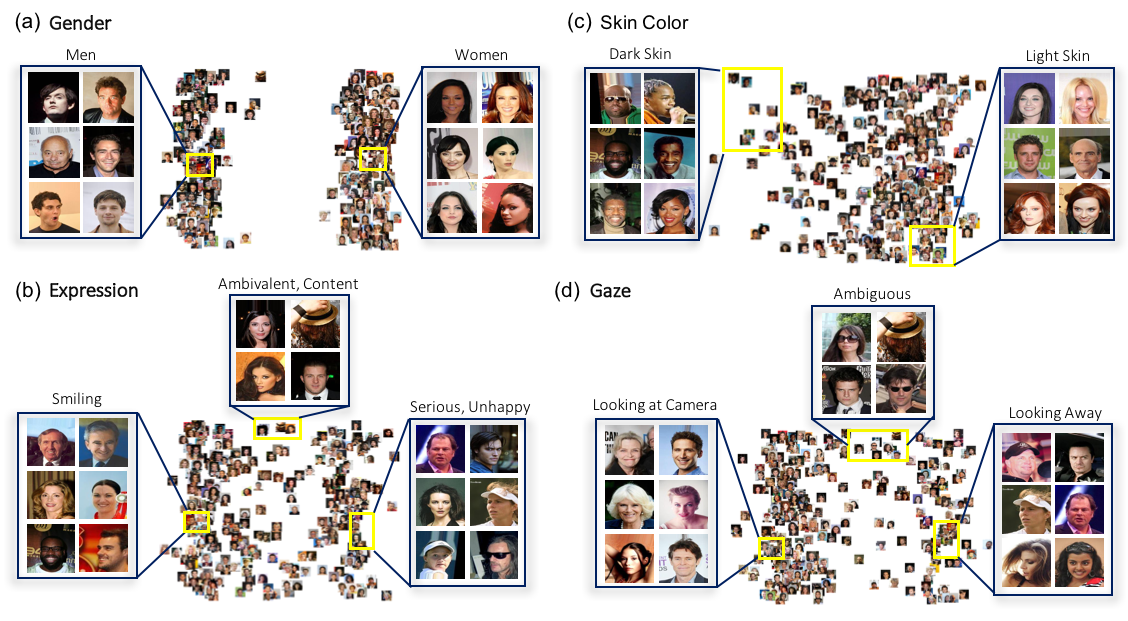}
\vspace{-5pt}
\caption{{\bf \textsc{CelebA} - Attribute specific embeddings.} Each plot shows one of the four different embedding dimensions produced by the CEN-mixture mode. The vertical axis in each subplot is randomly assigned for visualization purposes. We show representative images from the embeddings space in yellow boxes. We can see that the CEN learns to disentangle the attributes. }
\label{fig:celebemb}
\vspace{-10pt}
\end{figure*}

\begin{figure*}[h]
\centering
\includegraphics[width=0.9\textwidth]{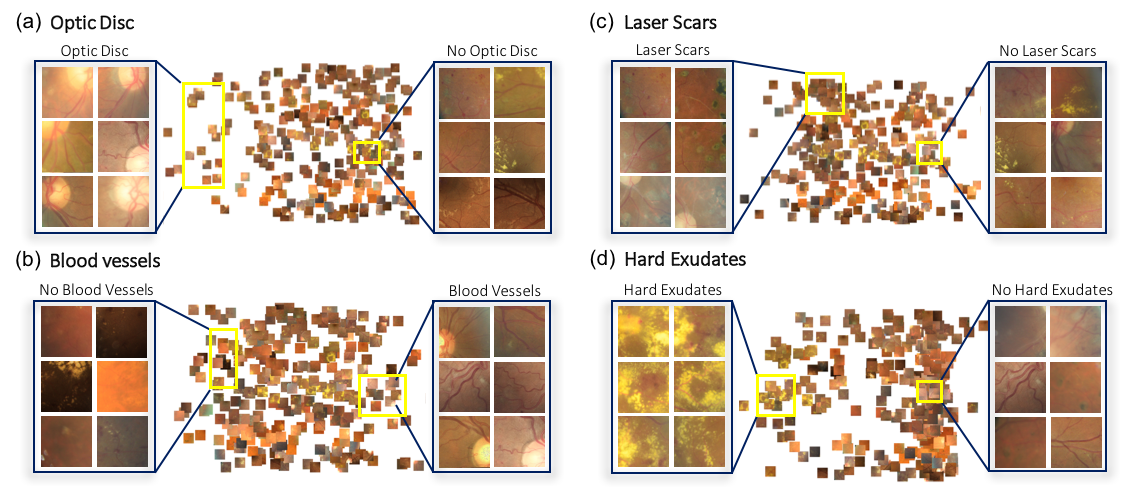}
\vspace{0pt}
\caption{{\bf \textsc{Retina} - Attribute specific embeddings.} Here we show a subset of four of the ten embeddings dimensions produced by the CEN-mixture model for the \textsc{Retina} dataset. Dimensions correlated well with visual features of diabetic retinopathy.}
\label{fig:retinaemb}
\vspace{-10pt}
\end{figure*}

\begin{figure}[h]
\centering
\includegraphics[width=0.45\textwidth]{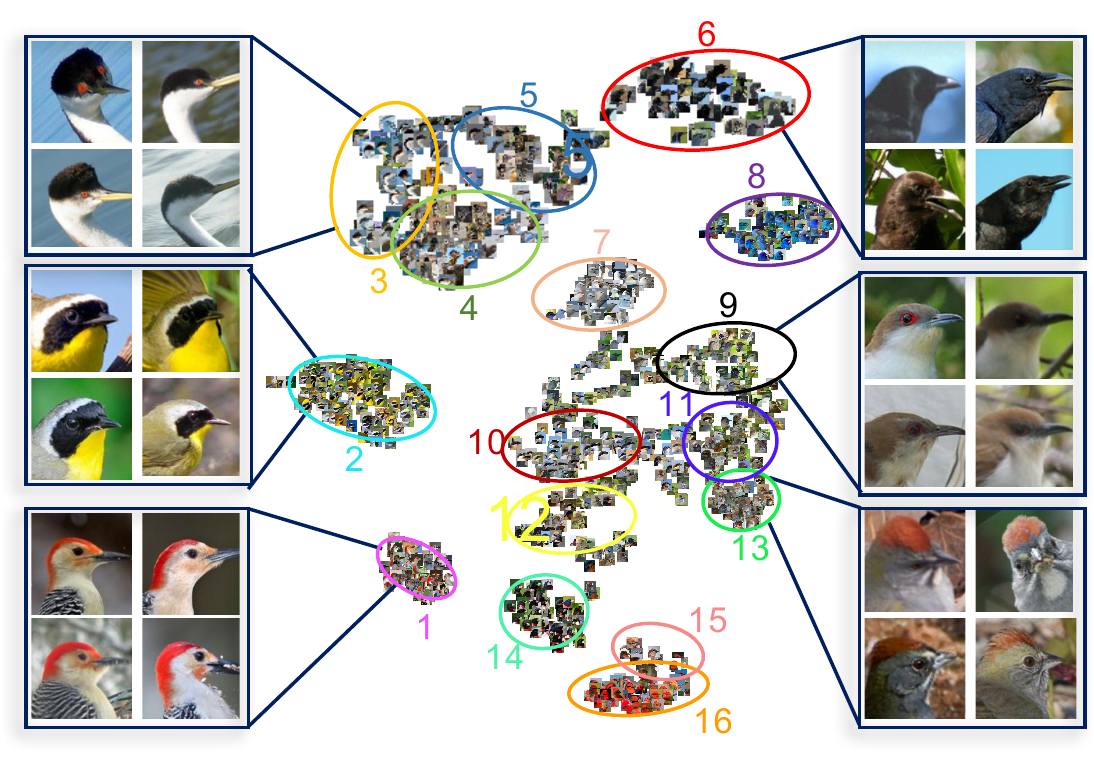}
\vspace{0pt}
\caption{{\bf \textsc{Birds} - t-SNE embedding.} Here we show a t-SNE \cite{maaten2008visualizing} plot of the four dimensional embedding produced by the full CEN model for the \textsc{Birds} dataset. Indexed ellipses are centered at the Gaussian mean of different ground truth species. Clusters correlated well with ground truth species of birds.}
\label{fig:birdsemb}
\vspace{-10pt}
\end{figure}

\subsection{Visualizing Disentangled Attributes}
Fig. \ref{fig:celebemb} shows the attribute specific embeddings of the four subspaces learned by the CEN for the \textsc{CelebA} dataset.
Fig. \ref{fig:celebemb}(a) shows that the embedding clearly separates the images according to \textit{gender}. On the very left of the \textit{expression} subspace (Fig. \ref{fig:celebemb}(b)) we can see that people are smiling with teeth showing while on the right they show serious or unhappy expressions. In the middle we see ambiguous expressions. Fig. \ref{fig:celebemb}(c) shows the subspace embedded along the \textit{skin color} attribute. On the two ends we see darker skinned and lighter skinned people. Fig. \ref{fig:celebemb}(d) shows the subspace for \textit{gaze direction} of people, showing people that are either looking at the camera or away from it. Again, in the middle we see people wearing sunglasses or looking in ambiguous directions making it difficult to assess their gaze direction. 

In Fig. \ref{fig:retinaemb} we show attribute specific embeddings learned for the \textsc{Retina} dataset in which no supervision was given to the workers for which attributes to pay attention to.
Here we select four dimensions that are most highly activated from the learned ten dimensional embedding vector. 
Other attribute dimensions attain trivial activations. 
This shows that our CEN is robust to value of $K$ (please see supplementary material for robustness analysis of $K$). 
Fig. \ref{fig:retinaemb}(a) shows the first dimension seemingly showing the presence or absence of the optic disc, a key feature of the retina. 
Fig. \ref{fig:retinaemb}(b) shows the subspace which discriminates between patches with blood vessels present and those without. 
Blood vessels are mostly concentrated and visually prominent around the optic disc, meaning that the two attributes are highly correlated. 
Regardless, our CEN is capable of distinguishing between the two attributes, as we see that images displaying blood vessels without optic discs are correctly embedded in Fig. \ref{fig:retinaemb}(a). 
Fig. \ref{fig:retinaemb}(c) plots the attribute that groups laser scars (named after consulting with an ophthalmologist)  and Fig. \ref{fig:retinaemb}(d) groups hard exudates, a key indicator for diagnosing diabetic retinopathy \cite{kalviainen2007diaretdb1}. 
A comparison of embedding qualities between baselines are presented in the supplementary material. 

Fig. \ref{fig:birdsemb} shows a t-SNE plot of the four dimensional embedding space learned by the CEN for the \textsc{Birds} dataset. Each ellipse center corresponds to the mean of a Gaussian distribution fit to the embedding coordinates for each ground truth species. We observe 16 compact clusters that directly correlate to the 16 ground truth species. Please refer to the supplementary materials for confusion plots of the ground truth species vs embedding clusters.

\begin{figure}[h]
\centering
\includegraphics[width=0.48\textwidth]{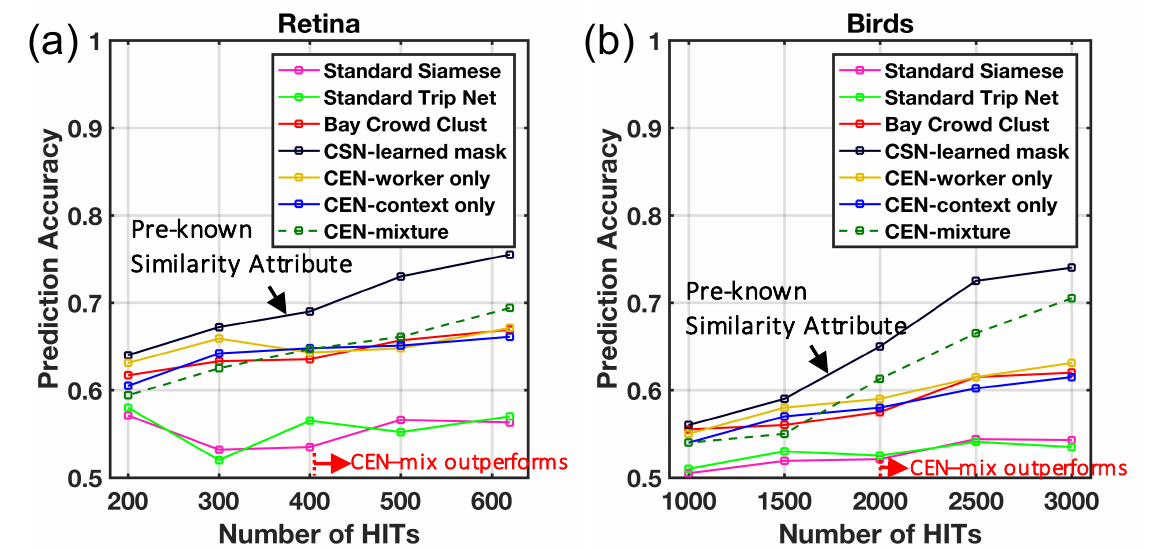}
\vspace{-10pt}
\caption{{\bf Held-out label prediction.} Prediction accuracy on held out labels for the \textsc{Retina} and \textsc{Birds} datasets plotted against the amount of available data during training.}
\label{fig:heldout_acc}
\vspace{-10pt}
\end{figure}

\subsection{Performance on Held-out Label Prediction}
Here we quantify the generalization performance of the baseline methods on held-out pairwise label predictions while varying the amount of training data. 
We measure the accuracy of the various model's predictions on the similarity estimates for an unseen grid clustered by a known worker. 
For a grid input $g$, worker input $w$, and image pair $i, j$, the model predicts $i$ and $j$ to be in the same group if $d < (\xi_{n} + \xi_{p}) / 2$. 
The test set is made up of 15\% of the dataset and consists only of entire grids that were not present in the training set.
This allows us to measure how well our CEN generalizes to new sets of images. 

Fig. \ref{fig:heldout_acc}(a) shows results for the \textsc{Retina} dataset. Standard Siamese Networks and Triplet Networks fail to capture the multiple attributes used to cluster the images and have the lowest prediction accuracy of $58.1\%$ and $58.5\%$. The Bayesian Crowd Clustering model, CEN worker, and CEN grid only models attain similar prediction accuracies of $67\%$. For the more challenging \textsc{Retina} dataset workers found it difficult to discover various attributes to cluster on and thus often fixated on a single attribute on all their HITs. However, we still benefit from modeling the context as the CEN-mixture model achieves prediction accuracy of $69.4\%$. 
The CSN model with learned masks obtains the highest accuracy of $75.5\%$, but it is important to note that this model was trained on triplets pre-labeled with the true similarity attributes used to cluster them. 

Fig. \ref{fig:heldout_acc}(b) shows the pairwise prediction accuracy for each model plotted against a varying number of training samples for the \textsc{Birds} dataset. The Bayesian Crowd Clustering model, CEN worker, and CEN grid only models attain similar prediction accuracies of $62\%$. The CEN-mixture substantially outperforms all baselines with a prediction accuracy of $70.5\%$ which is only $3.5\%$ below the accuracy of the CSN model which uses ground truth labels.

\subsection{Image Grid Synthesis}
\begin{figure}[t]
\centering
\includegraphics[width=0.32\textwidth]{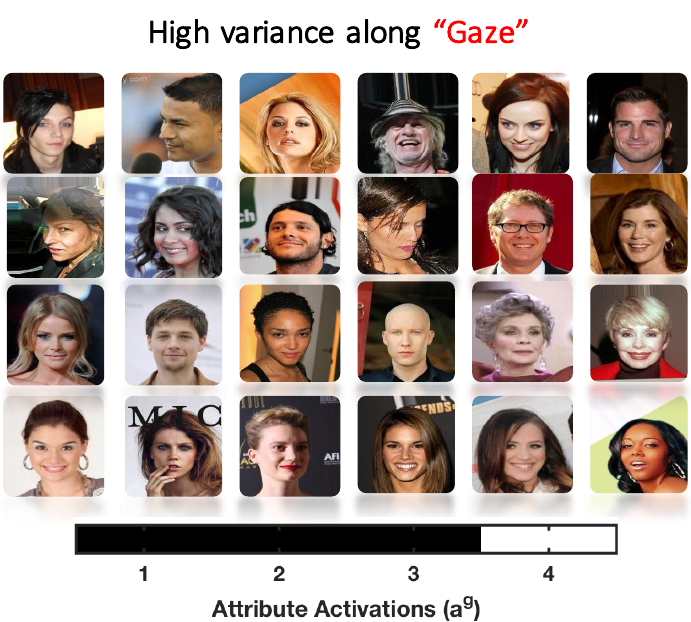}
\vspace{0pt}
\caption{{\bf Synthesized image grids.} Our context encoder can be used to generate collections of images that highlight specific attributes. The shown grid has high variance along the gaze direction attribute and low variance for the others.} 
\label{fig:gridmod}
\vspace{-10pt}
\end{figure}

Being able to synthesize image grids that highlight specific attributes may be useful in active learning  where the data collector seeks to obtain similarity estimates along particular visual attributes. 
We randomly generate ten million image grids and individually pass them through the context encoder and extract the grid attribute activation vectors $a^{g}$ for each grid. 
We take a softmax activation over the $a^{g}$s and select grids that have low entropy, thus choosing grids that are highly expressive for a particular attribute. 
Fig. \ref{fig:gridmod} shows a generated grid with the lowest entropy for the gaze attribute. 
We see low variance among the images along other attributes such as gender and skin color, while there is high variance for `gaze'. 
This suggests that in order for a grid to emphasize a particular attribute, the contained items should be similar in all but one high variance attribute. 

\vspace{-5pt}
\section{Conclusion} \label{sec:conclusions}
\vspace{-5pt}
We proposed a novel deep neural network that jointly learns attribute specific embeddings, worker models, and grid context models from the crowd. 
By comparing to several baseline methods, we show that our model more accurately predicts the attributes used by individual workers and as a result produces better quality image embeddings.

In future we plan to incorporate relative similarity estimates and the learning of representations directly from images \cite{veitconditional,schroff2015facenet}.
Although currently we model each worker individually, in practice there may be similarity between different workers that could be discovered through clustering \cite{kajino2013clustering}.
Finally, our grid context encoder enables us to generate sets of images that highlight specific attributes. 
By combining this with active learning we can potentially speed up the collection of annotations from the crowd \cite{tamuz2011adaptively,liang2014beyond}.

\vspace{-15pt}
{\small \paragraph{Acknowledgements} We thank Google for supporting the Visipedia project and AWS Research Credits for their donation.}






\bibliography{context_net}
\bibliographystyle{ieee}

\setlength{\itemindent}{-.5in}

\newcommand{\beginsupplement}{%
		\setcounter{section}{0}
   		\renewcommand\thesubsection{\Alph{subsection}}
        \setcounter{table}{0}
        \renewcommand{\thetable}{S\arabic{table}}%
        \setcounter{figure}{0}
        \renewcommand{\thefigure}{S\arabic{figure}}%
     }

\beginsupplement

\section*{Supplementary Material} \label{sec:results}
Here we provide the model architecture and training details for our CEN models along with additional experimental results.

\subsection{Model Parameters} 
For both datasets, the worker and context encoders are fully connected neural networks consisting of two hidden layers each with $200$ neurons with ReLU activations. The image encoder has one hidden layer with $200$ units and outputs a $K$ dimensional embedding vector. For the \textsc{CelebA} dataset, the embedding dimension $K$ is set to four since we provide the workers with four different attributes to cluster on. For the \textsc{Retina} dataset, we set $K = 10$ since we do not know \textit{a priori} how many different attributes the workers will use. We jointly train the three encoders with a mini-batch size of $100$ using ADAM with $\alpha = 0.001, \beta_{1} = 0.9,$ and $\beta_{2} = 0.999$. We experimented with various learning rates $\alpha \in \{0.00001, 0.0001, 0.001, 0.01\}$ and found that the CEN performance was robust to these variations. The regularization constants are set to $\lambda_{1} = 5\textsc{e}-6$ and $\lambda_{2} = 0.001$. We experimented with $\lambda_{1} \in \{1\textsc{e}-6, 5\textsc{e}-6, 1\textsc{e}-5, 5\textsc{e}-5\}, \lambda_{2} \in \{0.0001, 0.0005, 0.001, 0.005, 0.01\}$ and saw that the  prediction accuracy decreases when $\lambda_{1} > 5\textsc{e}-6, \lambda_{2} > 0.001$, but relatively stable otherwise. Hence, we choose the largest possible learning rate to reduce training time.

\begin{table}[b]
\caption{{\bf Impact of positive similarity weight $\gamma$} on label prediction accuracy. $\gamma = 6$ was used for all results presented in the main paper.}
\label{table:accuracy}
\begin{center}
\begin{small}
\begin{tabular}{|l|l|l|l|l|l|}
\hline
& $\gamma = 1$ & $\gamma = 4$ & $\gamma = 6$ & $\gamma = 8$ & $\gamma = 10$ \\ \hline
\hline
\textsc{CelebA} & 68.5\% & 69.8\% & 69.8\% & 69.3\% & 69.2\%   \\ \hline
\textsc{Retina} & 68.1\% & 69.3\% & 69.4\% & 69.1\% & 68.7\%   \\ \hline
\end{tabular}
\end{small}
\end{center}
\label{table:gamma}
\end{table}

The positive margin, negative margin, and positive similarity weight are each set to $\xi_{p} = 1, \xi_{n} = 6$ and $\gamma = 5$, respectively. The prediction accuracy decreased when $\xi_{n}/\xi_{p} < 4$. We experimented with varying positive similarity weights $\gamma \in \{1, 2, ..., 10\}$ and found that $\gamma = \{4, 5, 6\}$ achieves similar best prediction accuracy when trained on the full dataset. In Table \ref{table:gamma} we show the impact of $\gamma$ on the label prediction accuracy for both datasets. The optimum value of $\gamma$ should be expected to change depending on the variance in the level of detail workers cluster grids. 
Models were trained for 20 epochs which we determined to be sufficient for learning interpretable embeddings. When utilizing all the data from 620 HITs, the training time was on average 2.5 minutes for both datasets running on CPUs (Macbook Pro 13-inch, Late 2012, 2.5 GHz Intel Core i5, 8GB RAM, Apple, CA, USA). 
Upon publication we will make the code for our GUI and CEN model available.

\begin{figure}[t!]
\centering
\includegraphics[width=0.5\textwidth]{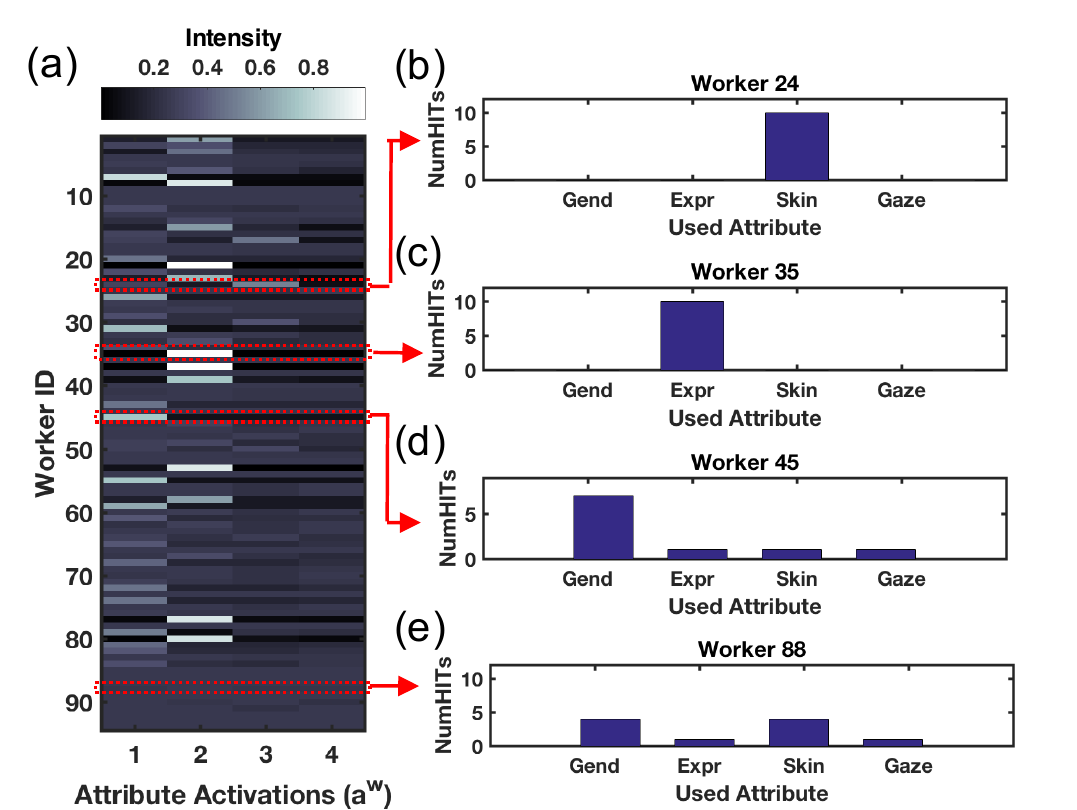}
\caption{{\bf Visualizing the worker model.} On the left we see the predicted attribute activation vectors for each worker from the \textsc{CelebA} dataset. Attribute dimension labels were inferred from worker annotations. Brighter colors indicate a stronger preference for a given attribute. On the right we show the actual attributes used by a set of representative workers inferred from their text annotations. We can see that our worker attribute predictions are consistent with the actual attributes used by the workers.} 
\label{fig:workermod}
\vspace{-15pt}
\end{figure}

\begin{figure*}[h]
\centering
\includegraphics[width=1\textwidth]{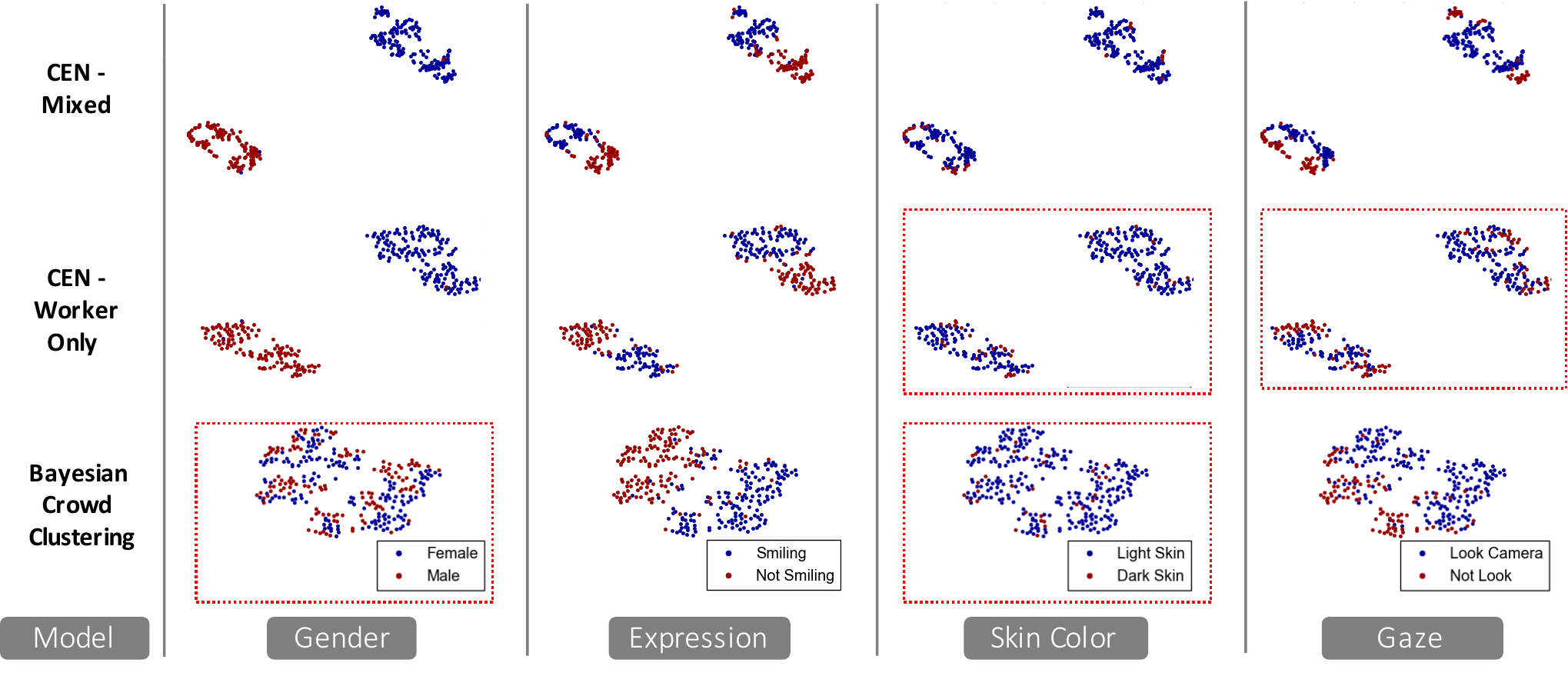}
\vspace{-15pt}
\caption{{\bf Comparing embedding quality.} 2D t-SNE projections of the four dimensional joint embedding space for three embedding models on the \textsc{CelebA} dataset. Colors denote binarized ground truth categories for each of the four attributes: gender, expression, skin color, and gaze direction. Dotted red boxes highlight attributes for which the CEN-mixture model produces more compact embeddings compared to the baselines.}
\label{fig:embcompare}
\vspace{-10pt}
\end{figure*}

\subsection{Interpretation of the Worker Model}
We explore the learned attribute activation vectors $a^{w}$ for each worker to examine if their prior biases were captured by the worker encoder. The output attribute activation vectors for each of the 94 workers are shown in Fig. \ref{fig:workermod}(a) as a stacked heatmap. On the right side of Fig. \ref{fig:workermod}, we show the distribution of attributes that four representative workers have used over the course of performing ten HITs, inferred from their text annotations. Fig. \ref{fig:workermod}(a) shows that our model predicts a high activation in $a^{w}_{3}$ for worker 24. In Fig. \ref{fig:workermod}(b) we can see that this worker consistently used the skin color attribute for all ten HITs they performed. This indicates that worker 24 had a strong prior bias towards grouping based on skin color and was unaffected by the different contexts formed by the grid. 
Note that it is highly unlikely that all ten randomly generated grids shown to worker 24  highlighted the skin color attribute. 
Fig. \ref{fig:workermod}(c) shows that worker 35 relied mainly on the expression attribute. Similarly for worker 45 we observe a strong bias towards the gender attribute as the worker encoder outputs a high activation for $a^{w}_{1}$. Worker 88 used a variety of attributes suggesting that they are more sensitive to the context provided by the grid. We observe a near uniform attribution activation vector for this worker. 

\begin{figure}[h]
\centering
\vspace{-10pt}
\includegraphics[width=0.35\textwidth]{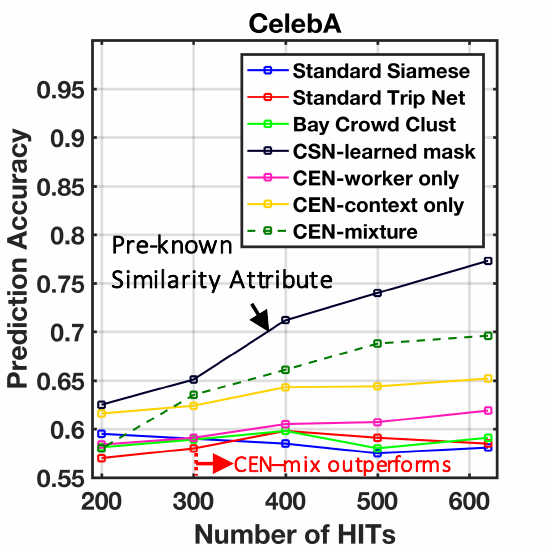}
\caption{{\bf Held-out Label Prediction on \textsc{CelebA}}. Prediction accuracy on held out labels for the \textsc{CelebA} dataset plotted against the amount of available data during training.
}
\vspace{-15pt}
\label{fig:facesingle}
\end{figure}

\begin{figure}[h]
\centering
\vspace{-10pt}
\includegraphics[width=0.35\textwidth]{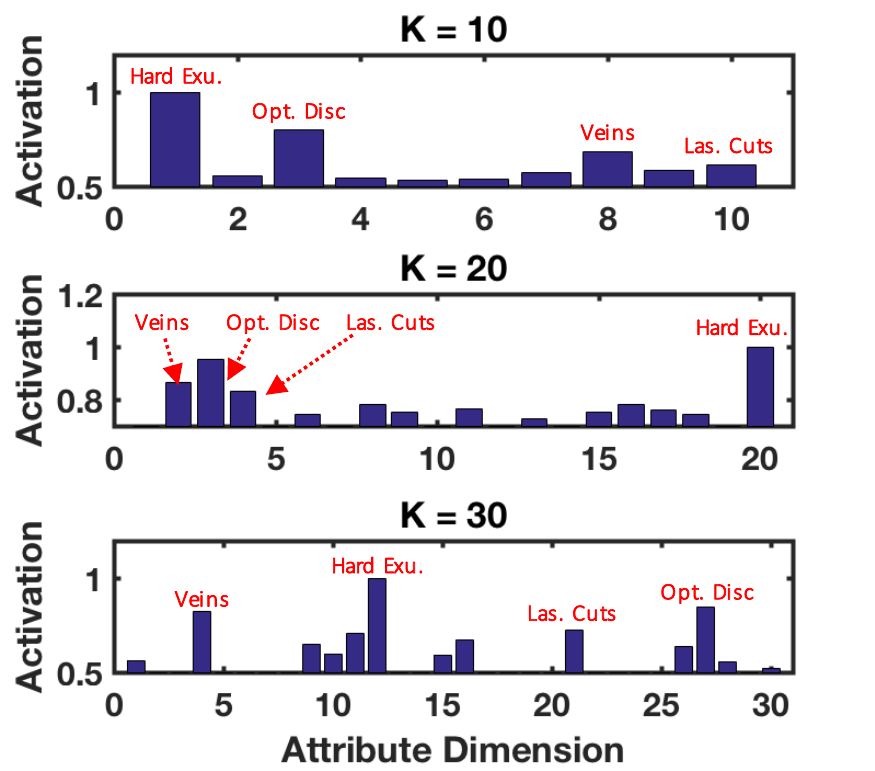}
\caption{{\bf Varying the embedding dimension for the \textsc{Retina} dataset}. We plot the average activations for each dimension of $a^{m}$ produced by the CEN-mixture model for three different values of embedding dimension $K$. Red labels were inferred from the text annotations. 
}
\label{fig:justifyk}
\vspace{-12pt}
\end{figure}

\subsection{Comparison of the Joint Embeddings}
We compare the quality of the embeddings for the \textsc{CelebA} dataset produced by the CEN-mixture model with those learned by baseline approaches that do not model context. We project the four dimensional joint embedding vectors $x_{i}$ down to two dimensions using t-SNE \cite{maaten2008visualizing} and color code each point according to its ground truth attribute. For each attribute, the ground truth categories were binarized for simplicity, i.e. smiling vs not smiling. In Fig. \ref{fig:embcompare} we show the low dimensional embeddings learned by the CEN-mixture model, CEN-worker only model, and the Bayesian Crowd Clustering baseline. The CEN-mixture model better separates the ground truth categories in the embedding space. This shows the positive impact of modeling context. The worker encoder only model finds well separated embeddings along the gender and expression attribute (which are relatively easy to distinguish) but does not perform well on the skin color and gaze attributes (which are attributes that workers more often disagree on). The Bayesian Crowd Clustering baseline has difficulty separating the gender and skin color attributes.

\subsection{Heldout Label Predictions on \textsc{CelebA}}
Fig. \ref{fig:facesingle}(a) shows the pairwise prediction accuracy for each model plotted against a varying number of training samples for the \textsc{CelebA} dataset. Standard Siamese Networks and Triplet Networks fail to capture the multiple attributes used to cluster the images and have the lowest prediction accuracy of $58.1\%$ and $58.5\%$. The Bayesian Crowd Clustering method slightly improves on that with an accuracy $59.1\%$. The worker only variant of our model achieves a prediction accuracy of $62.1\%$. This is superior to Siamese Networks and Bayesian Crowd Clustering but still fails to capture the tendency of workers to shift their clustering criterion based on the context highlighted by images in the grid. 
The context only model variant performs substantially better with a prediction accuracy of $65.2\%$. This indicates that the context information is indeed influencing the worker's decisions. Finally, the CEN-mixture outperforms all previous baselines with a prediction accuracy of $69.8\%$ ($75.1\%$ when trained on noiseless labels). The CSN model with learned masks obtains the highest accuracy of $77.3\%$, but it is important to note that this model was trained on triplets pre-labeled with the true similarity attributes used to cluster them.  
The CEN-mixture model achieves strong predictive performance without any prior knowledge of the similarity attributes.  
 
\subsection{Prior Number of Attributes}
For the \textsc{retina} dataset, we do not know the number of number of attributes the workers will use. Hence, we set $K = 10$ which serves as our prior guess of an upperbound on the number of attributes the workers are going to use. Although the attribute vector dimension was set to $K = 10$, we observed that four dimensions were consistently highly activated across different values of $K$. In Fig. \ref{fig:justifyk} we see that the attribute dimensions we selected are the four most highly activated dimensions of $a^{m}$ for $K = 10, 20,$ and $30$. 

\begin{figure}[h]
\centering
\includegraphics[width=0.35\textwidth]{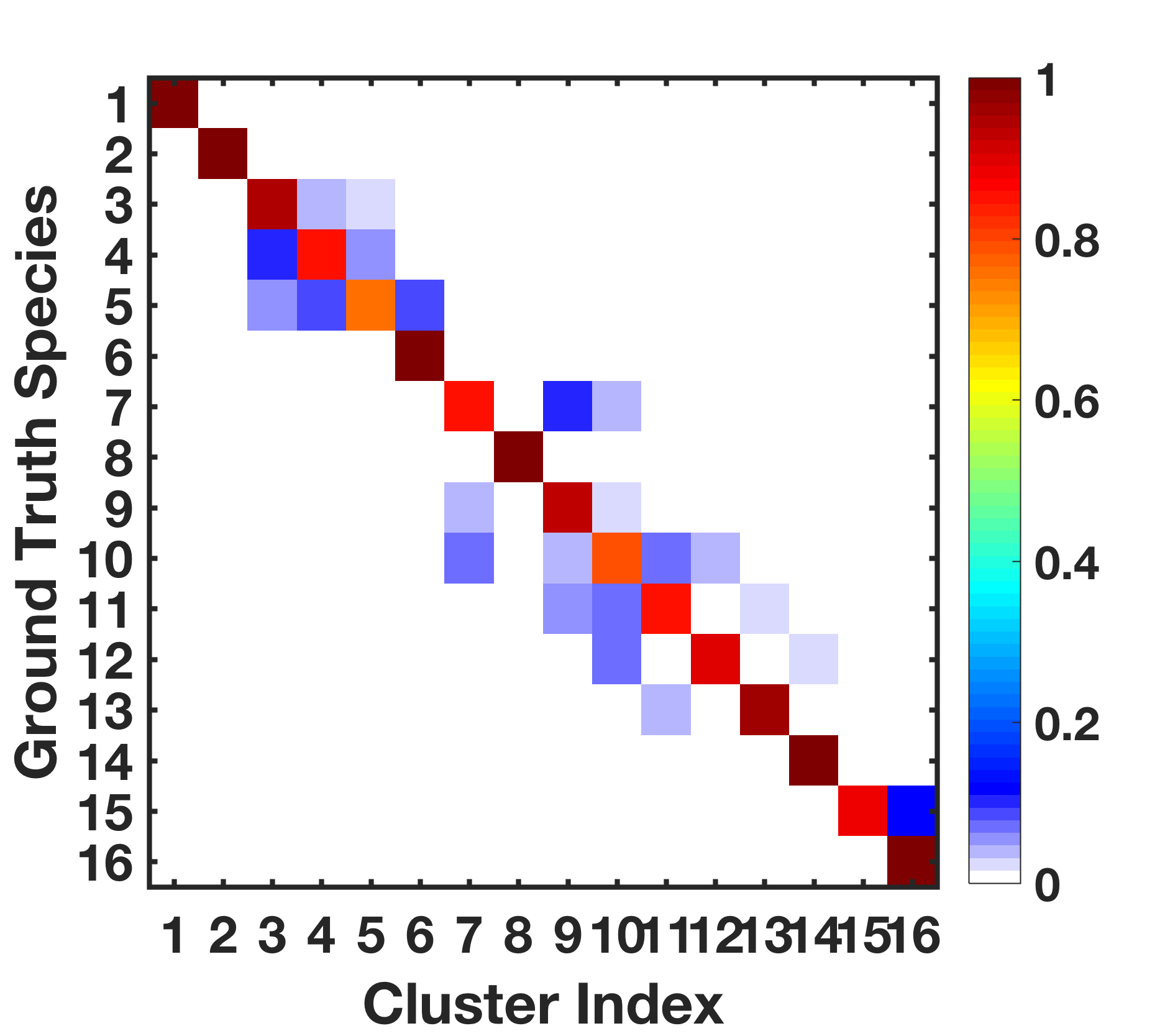}
\caption{{\bf Confusion plot for the \textsc{Birds} dataset}.}
\label{fig:birdsconfusion}
\vspace{-15pt}
\end{figure}

\subsection{Clustering Learned Embeddings}
For the \textsc{Birds} dataset, we perform K-means clustering on the learned 4 dimensional embedding space with $K = 16$ and compare the ground truth bird species of an image with its assigned cluster. To quantify the agreement between the ground truth species and the learned clusters, we use the multi-class version of Matthew's Correlation Coefficient (MCC) \cite{jurman2010mcc}, where $\textrm{MCC} = 1$ indicates perfect prediction, and a value between $-1$ and $0$ denotes total disagreement depending on the true distribution.   

\begin{equation}
\label{eq:mcc}
\textrm{MCC} = \frac{\sum\limits_{k,l,m=1}^N C_{kk}C_{ml} - C_{lk}C_{km}}{
\sqrt{\sum\limits_{k=1}^N \sum\limits_{l=1}^N C_{lk} \sum\limits_{\substack{f,g=1\\ f\not=k}}^N C_{gf}}
\sqrt{\sum\limits_{k=1}^N \sum\limits_{l=1}^N C_{kl} \sum\limits_{\substack{f,g=1\\ f\not=k}}^N C_{fg}}
}
\end{equation}

The confusion plot in Fig. \ref{fig:birdsconfusion} reveals high correlation ($\textrm{MCC} = 0.914$) between the ground truth species and the learned clusters, suggesting that the CEN is able to make fine-grained distinctions amongst bird species despite highly noisy training data ($25.6\%$ for the \textsc{Birds} dataset). 

To show that the learned embeddings are useful for fine-grained classification tasks, we trained a CNN with the cluster assignments as the image category labels and compared the resulting accuracy when training on the ground truth labels (900 images for training and 100 for testing). Our `embedding label' CNN resulted in a test accuracy of $68.1\%$, while the ground truth CNN produced an accuracy of $76.2\%$.

\subsection{Limitations}
If workers do not have a diverse set of abilities it will be challenging to learn embeddings that capture all subtle variations in a given dataset. However, in our experiments we observed that MTurkers discovered small distinctions in challenging domains e.g. retina images and bird species (see Fig. \ref{fig:retinaemb}, \ref{fig:birdsemb})

Our context model assumes that the ordering of the images in the grid does not effect clustering behavior. 
In practice this may have some effect on the workers.  
To produce disentangled attribute vectors we assume that a majority of the grids are clustered along a single attribute.
However, from our experiments we observe that this only has to be very weakly satisfied as many workers used a mixture of attributes.

\end{document}